\documentclass[10pt,letterpaper]{article}

\usepackage{upquote}
\usepackage{cogsci}
\usepackage{pslatex}
\usepackage{apacite}
\usepackage{csquotes}
\usepackage{graphicx}
\usepackage{verse}
\usepackage{import}
\usepackage{natbib}
\graphicspath{{images/}}

\title{A rational analysis of curiosity}
 
\author{{\large \bf Rachit Dubey (rach0012@berkeley.edu)} \\
  Department of Education, University of California at Berkeley, CA, USA
  \AND {\large \bf Thomas L. Griffiths (tom\_griffiths@berkeley.edu)} \\
  Department of Psychology, University of California at Berkeley, CA, USA}

\begin{document}

\maketitle

\begin{abstract}
We present a rational analysis of curiosity, proposing that people's curiosity is driven by seeking stimuli that maximize their ability to make appropriate responses in the future. This perspective offers a way to unify previous theories of curiosity into a single framework. Experimental results confirm our model's predictions, showing how the relationship between curiosity and confidence can change significantly depending on the nature of the environment. 

\end{abstract}

\section{Introduction}

In 1928, upon returning from a vacation, Alexander Fleming, who was a professor of Bacteriology at St. Mary's Hospital in London, noticed how a mold floating in one of his dirty petri dishes held the surrounding bacteria at bay. This peculiar event led him to develop a hypothesis that would be a prelude to the development of penicillin. The history of science abounds with incidents in which an event piqued the curiosity of a scientist thereby leading to important discoveries (other examples include Curie, Faraday, and Planck). For this reason, intellectual curiosity has long been recognized as the essence of science. In fact, Herbert Simon famously titled a 1992 talk given at Carnegie Mellon as `The cat that curiosity couldn't kill' and described curiosity to be not only the beginning of all science, but also its end (Gobet, 2016).

Considering how important curiosity is to scientific discoveries and many other aspects of cognition, it is surprising that our understanding of curiosity as a psychological phenomenon remains quite limited (Simon, 2001; Gottlieb et al., 2013; Kidd \& Hayden 2015). Encouragingly, the field has seen a revived interest in curiosity in recent years with psychologists and neuroscientists beginning concentrated efforts to study curiosity systematically (Kang et al, 2009; Gruber, Gelman, \& Ranganath, 2013; Law et al., 2016; Walin, O'Grady \& Xu, 2016). However, much previous work on curiosity has either focused on defining a taxonomy for curiosity or providing a mechanistic explanation of curiosity (Kidd \& Hayden, 2015). This means that while we have made some progress describing the psychological processes involved in human curiosity, we have not satisfactorily provided an explanation of the purpose and function of curiosity. For example, a commonly held notion about the function of curiosity is that it motivates learning (Loewenstein, 1994; Kidd \& Hayden, 2015). Although it is easy to say that learning is the goal of curiosity, this is not very precise in its meaning. How does curiosity facilitate learning? Why does it do so?  

In light of this, in this paper we present a rational analysis of curiosity in the spirit of Anderson (1990) and Marr (1982) with the goal of providing  a purposive explanation of curiosity. Our work shows that a rational analysis can predict many aspects of curiosity without making assumptions about its mechanisms. We start by defining an abstract representation of the problem that curiosity solves and making a small number of assumptions about the nature of the environment. Following that, we explore the optimal solution to this problem in light of these assumptions. Our theory posits that people are curious about stimuli that maximally increase the usefulness of their current knowledge. Depending on the structure of the environment, the stimuli that maximize this value can either be ones that are completely novel or that are of intermediate complexity. As a consequence, our rational analysis provides a way to unite previously distinct theories of curiosity into a single framework. 

The rest of the paper is structured as follows. We first review previous theories of curiosity and then introduce our rational model of curiosity. Following that, we show how our model offers support to previous distinct accounts of curiosity thereby unifying them in a single model. We then conduct a behavioral experiment to test our model's predictions and evaluate how our model accords with human curiosity. We close with a discussion of the implications of our results.

\section{Models of curiosity}

A number of theories have been proposed in the past to describe the psychological processes involving curiosity. In this section, we describe these theories in brief and provide their individual strengths and weaknesses. 


\textbf{Curiosity based on novelty. }Several psychological theories have linked curiosity with novelty by hypothesizing that gaining information about novel stimuli is intrinsically rewarding (Berlyne, 1950; Smock \& Holt, 1962). Berlyne (1960) called this ``perceptual curiosity" and described it as a driving force that motivates an organism to seek out novel stimuli which diminishes with an increase in exposure. This has also been supported by some neuroscientific studies that show that novel stimuli activate reward-responsive areas in the brain (Ranganath and Rainer, 2003; D{\"u}zel et al., 2010). However, a severe limitation of this theory is that it assumes that it is optimal for an individual to explore novel stimuli in all environments. A novel stimulus doesn't necessarily mean that it contains information that is useful or generalizable to an individual. This is also pointed out by previous studies that show that exploration based only on novelty could lead agents to be trapped in unlearnable situations (Gottlieb et al., 2013). 

\textbf{Curiosity based on information-gap. }One of the most popular theories of curiosity is the information-gap hypothesis proposed by Loewenstein (1994). According to the information-gap hypothesis, curiosity arises whenever an individual has a gap in information prompting it to complete its knowledge and resolve the uncertainty. Thus, curiosity peaks when one has a small amount of knowledge but it diminishes when one knows too little or too much about the stimuli. A number of studies have supported this prediction and showed that curiosity is an inverted U-shaped function of confidence, with people showing the highest curiosity for topics that they were moderately confident about (Kang et al., 2009; Baranes, Oudeyer, \& Gottlieb, 2015). Berlyne (1960) called this form of curiosity ``epistemic curiosity" and described it as a drive to acquire knowledge. While this theory has considerable strengths, it is also constrained in that an individual can only be curious about stimuli in known contexts. Thus, if an individual has no prior knowledge about stimuli in the environment then it is not clear how curiosity will function in that environment (as one will not be curious about any stimuli in an environment that it has no prior knowledge of). 

\textbf{Curiosity based on learning progress. }A third theory concerning curiosity is guided by the hypothesis that learning progress generates intrinsic reward (Schmidhuber,1991; Schmidhuber, 2010). This hypothesis proposes that the brain is intrinsically motivated to pursue tasks in which one's predictions are always improving. Thus, an individual will not be interested in tasks that are too easy or too difficult to predict but will rather focus on tasks that are learnable. Based on Schmidhuber's framework, a number of papers in developmental robotics have supported this idea showing how an agent can explore in an unknown environment (Oudeyer \& Kaplan, 2006; Oudeyer, Kaplan, \& Hafner, 2007). While this theory can probably describe some forms of curiosity, it is again constrained in explaining curiosity in certain environments. For example, if an agent is ever present in an environment that has many difficult tasks then it is not clear how curiosity will work (as an agent will not be curious about anything within that environment). 

\textbf{Summary and prospectus. }While each of these theories have their strengths, we first note that all the above theories are concerned with describing how curiosity functions and how it relates to different psychological factors. However, none of these theories satisfactorily provide an explanation as to \textit{why} curiosity works the way it does. Second, we believe that all these theories need not be in contention but are all rather special cases of curiosity. As we will describe in the rest of the paper, our rational model supports each of these theories and unifies them in one common model.

\section{Rational model of curiosity}

In this section, we detail our rational model of curiosity. We first consider the abstract computational problem underlying curiosity and then formally derive an optimal solution to this problem.

\subsection{Computational problem underlying curiosity}

Suppose that an agent is in an environment with $n$ stimuli, each of which provides a reward if the appropriate response is produced. The goal of the agent is to decide what to explore in the environment in order to maximize its knowledge and hence maximize rewards in the future.

The environment determines the  probability with which each stimulus occurs in the environment. Let $p_k$ denote the ``need probability" that a stimulus $k$ will occur in the future (Anderson, 1990). Given this, the agent assigns a confidence value $c_k$ to each stimulus in the environment. $c_k$ denotes the probability the agent knows the correct response to the $k$th stimulus. This probability increases at a decreasing rate with respect to the number of exposures $h_k$ with that stimulus. $h_k$ denotes the number of times the agent has been exposed to the $k$th stimulus. For convenience, we describe the relationship between $c$ and $h$ by a bounded growth function, 

\begin{equation}{c_k = 1-e^{-h_k}}. \end{equation}
However, our predictions hold for any monotonically increasing function. 

Next, the agent computes the value of its overall knowledge. The value, denoted as $V$, is a function of the need probability $p$ and the confidence factor $c$ and is given as follows:

\begin{equation}{V = \displaystyle\sum_{k}p_k.c_k.} \end{equation}
According to this equation, the value of an agent's knowledge is simply the chance of successfully responding to the next stimulus computed by summing over all stimuli in the environment. 

The goal of the agent is to increase the value of its current knowledge $V$, which it can do so by taking actions to increase $h$ for the various stimuli in the environment. So the computational problem that the agent has to solve is choosing which stimulus to explore further i.e. deciding which stimulus $k$ to increase $h_k$ for.

\subsection{Deriving an optimal solution}

To solve the problem of choosing which stimulus to explore further, the agent can evaluate the change in $V$ as it explores each stimulus in the environment. Thus for every stimulus $k$ in the environment, the agent should compute the change in its knowledge that would result from exploring that stimulus. The stimulus that causes the largest increase in the overall value should be explored first. This computation can be done simply by  differentiating $V$ with respect to $h_k$,

\begin{equation}{\frac{\mathrm{d}V}{\mathrm{d}h_k} = p_k.\frac{\mathrm{d}c_k}{\mathrm{d}h_k}}. \end{equation}

An agent operating according to this model will explore stimuli that have a high rate of change of value w.r.t  exposure associated with them. This rate of change, i.e. $\frac{\mathrm{d}V}{\mathrm{d}h_k}$, is simply the curiosity the agent has for knowing the $k$th item, which we denote as $\Omega_k$. The agent will explore stimuli in the environment that it is most curious about. In this way, curiosity helps the agent to achieve its goal of maximizing its knowledge. 

Under the choice of the form of $c_k$ given in Equation 1, we calculate this derivative as follows:

\begin{equation}{\Omega_k = p_k.\frac{\mathrm{d}(1-e^{-h_k})}{\mathrm{d}h_k}}. \end{equation}
Upon differentiation, we get the relationship of curiosity $\Omega_k$, with respect to need probability $p_k$ and exposure $h_k$ as follows: 

\begin{equation}{\Omega_k = p_k.e^{-h_k}}. \end{equation}

\subsection{Relationship to previous models}

\begin{figure}
    \centering
    \includegraphics[scale = 0.18]{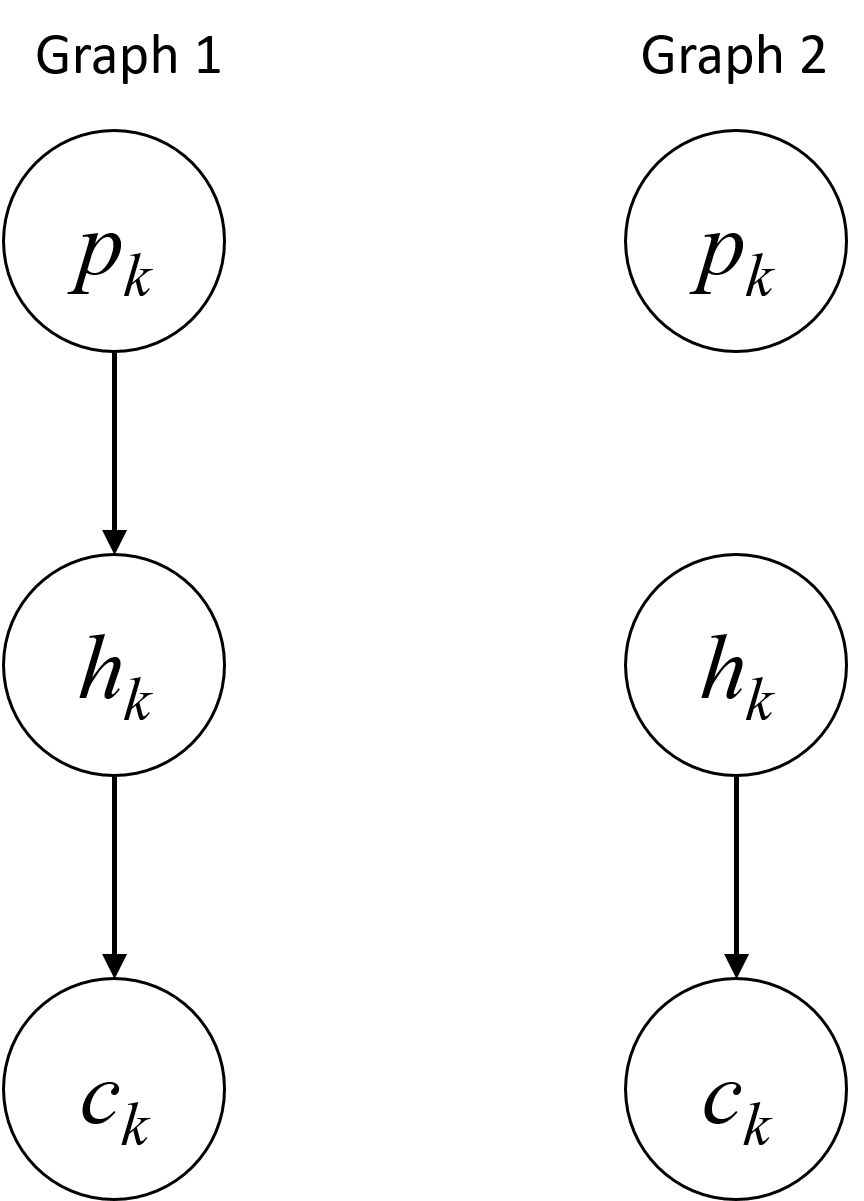}
    \caption{Relationship between need probability $p_k$, exposure $h_k$, and confidence $c_k$ in different environmental structures. Graph 1 shows an environment in which need probability is related to exposure and subsequently confidence. Graph 2 shows an environment in which need probability is independent of exposure and confidence.}
\end{figure}

Having a formal account of curiosity, we now describe how our model relates to previous theories of curiosity. 

First, we note that in our rational model framework, two different forms of environmental structure can exist. The first form comes in when the agent is an environment where $p_k$ is related to $h_k$ (as described in Graph 1, Figure 1). In this environment, stimuli frequently encountered by an agent are more likely to be needed in the future. Thus, the probability that the agent will require a stimulus in the future determines the number of times the agent is exposed to the stimulus which in turn determines the confidence of the agent in knowing that stimulus. The second form comes in when the agent is in an environment where $p_k$ and $h_k$ are independent of each other (as described in Graph 2, Figure 1). In this environment, the agent can encounter any stimulus in the future regardless of their previous occurrence.  

\textbf{Novelty based curiosity. } According to theories that are based on curiosity driven by novelty, an agent is most curious about stimuli that it is least confident about:

\begin{equation}{\Omega_k = 1-c_k}. \end{equation}

According to our rational model, when the agent is in
an environment where $p_k$ and $h_k$ are independent of each other (as described in Graph 2, Figure 1), the relationship between curiosity and exposure will be the one described in Equation 5 where $p_k$ is simply a constant value. Thus, curiosity is highest when exposure is lowest and it decreases as exposure increases i.e. curiosity is highest for novel stimuli. The relationship between curiosity and confidence can be rewritten as

\begin{equation}{\Omega_k = p_k.(1-c_k).} \end{equation}
If $p_k$ is equal for all $k$, this reduces to Equation 6. Thus, when need probability and exposure are not related to each other, our rational model is similar to the previously proposed novelty based curiosity theory. 

\textbf{Information-gap hypothesis.} When the agent is in an environment $p_k$ and $h_k$ are related to each other (as in Graph 1, Figure 1), then $p_k$ is proportional to $h_k$ and the relationship between curiosity and exposure given in Equation 5 reduces to   

\begin{equation}{\Omega_k \propto h_k.e^{-h_k}}. \end{equation}
Subsequently, using Equation 1, confidence will be related to curiosity as

\begin{equation}{\Omega_k \propto -\log(1-c_k).(1-c_k)}. \end{equation}

Interestingly, this relationship between curiosity and confidence is highly similar to the one described by the information gap hypothesis. Loewenstein used Shannon's (1948) entropy formula to describe the relationship between curiosity and confidence as below:

\begin{equation}{\Omega_k = -\log(c_k).(c_k)}. \end{equation}

Both the information-gap theory and our model predict that an inverted U-shape relationship exists between curiosity and confidence. In this view, when an agent exists in an environment where need probability is related to exposure, our rational model relates to the information-gap theory.

\textbf{Relationship to curiosity based on learning progress. } According to the learning progress hypothesis, an agent is intrinsically motivated to pursue tasks in which predictions are constantly improving thereby avoiding boring or extremely complicated tasks. An agent operating under this model ends up exploring stimuli of ``intermediate complexity". 

Our model proposes that an agent will explore stimuli that maximize the value of its current knowledge. In an environment where need probability and exposure are related to each other, then curiosity is highest for stimulus with moderate exposure i.e. intermediate complexity (Equation 8 and 9). Thus, in this environment, an agent that aims to maximize its knowledge behaves similarly to an agent whose curiosity is driven by learning progress. 
 
\textbf{Summary. } Whereas previous theories associated curiosity to factors such as novelty, knowledge gap, and learnability, our model shows that depending on the structure of the environment, an agent's curiosity can be driven by any of these factors. In this way, our rational model allows to bridge previous theories related to curiosity in a single framework. In an environment where need probability and exposure are related, our rational model associates with the information gap and learning progress hypothesis. In an environment where need probability and exposure are not related, our model is akin to the novelty-based theory of curiosity. 
 
\subsection{Empirical predictions}

\begin{figure}[t]
    \centering
    \includegraphics[scale = 0.35]{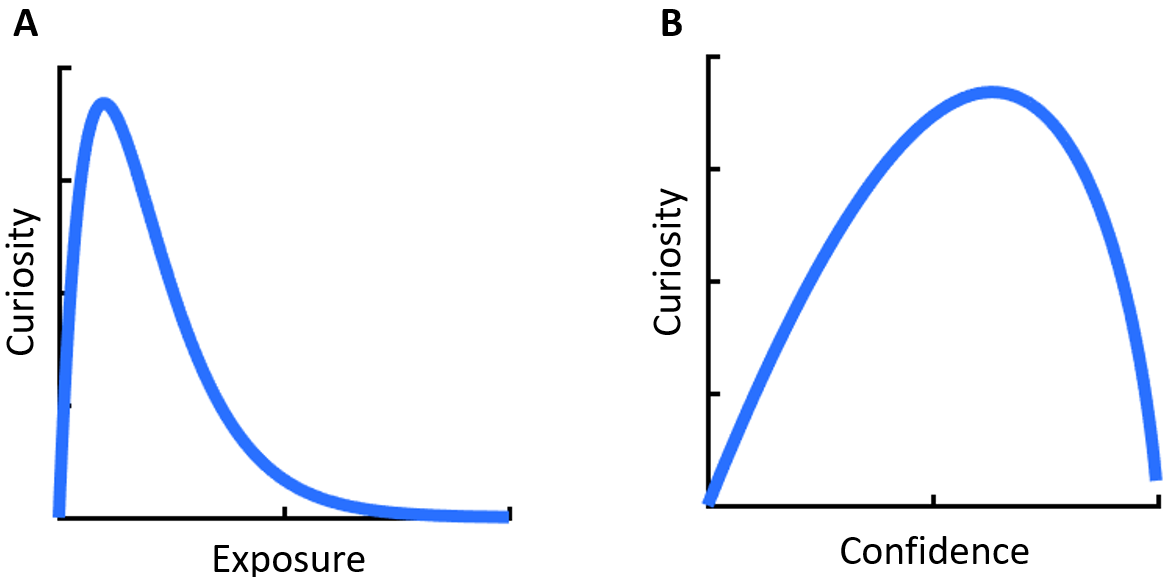}
    \caption{Relationship between a) curiosity and exposure, and b) curiosity and confidence in an environment where need probability is related to exposure (Graph 1, Figure 1).}
\end{figure}

\begin{figure}[b]
    \centering
    \includegraphics[scale = 0.35]{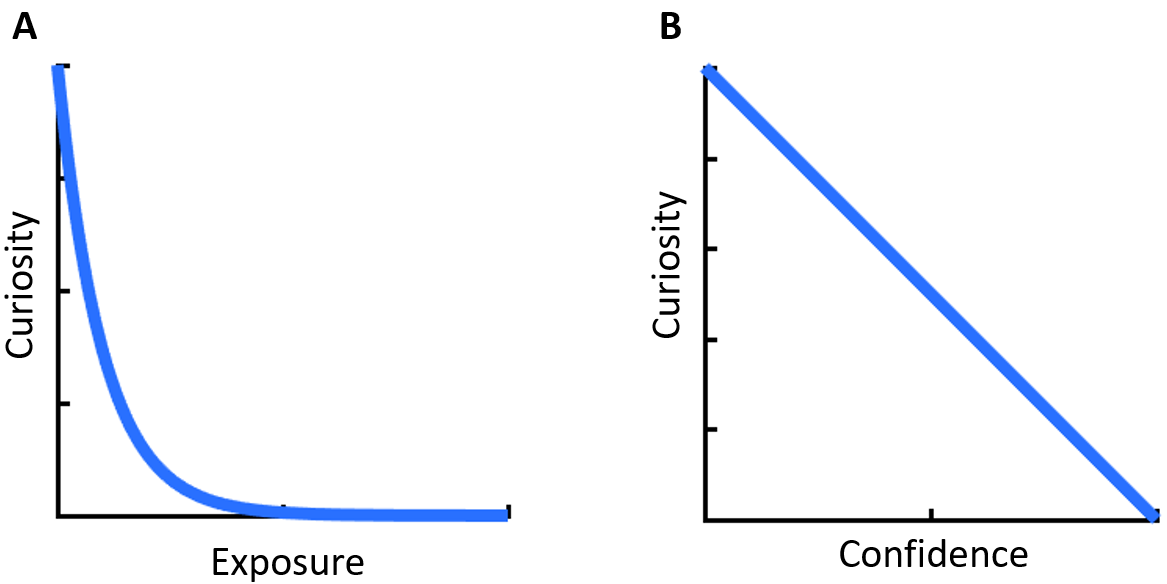}
    \caption{Relationship between a) curiosity and exposure, and b) curiosity and confidence in an environment where need probability is independent of exposure (Graph 2, Figure 1).}
\end{figure}

The rational model presented above makes two different empirical predictions.

\textbf{Prediction 1. }The first prediction arises when the agent is in an environment where the relationship between $p_k$ and $h_k$ holds true (as in Graph 1, Figure 1). The relationship between curiosity and exposure can be described using Equation 8 and between curiosity and confidence using Equation 9. Thus, Equation 8 predicts that an inverted U-shape relationship will exist between curiosity and exposure and Equation 9 similarly predicts that curiosity will be highest when the agent is moderately confident about a stimulus (see Figure 2). We test this prediction in the \textit{confidence} sampling condition of our behavioral experiment. 

Note that this prediction fits the information gap and learning progress hypothesis which also predict an inverted U-shape curve between curiosity and confidence. While several studies have supported the existence of this U-shaped relationship, our model also predicts how to make this effect go away as described in our second prediction.  

\textbf{Prediction 2. }Our second prediction comes in when the agent is in an environment where the relationship between  $p_k$ and $h_k$ no longer holds true (as described in Graph 2, Figure 1). Then the relationship between curiosity and exposure will be the one described in Equation 5 and the relationship between curiosity and confidence will be that given in Equation 7. Equation 5 predicts that curiosity is highest when exposure is lowest and it decreases as exposure increases. Similarly, Equation 7 predicts that curiosity will be highest when confidence is the lowest (also shown in Figure 3). We test this prediction in the \textit{uniform} sampling condition of the behavioral experiment.

While this prediction accords with the prediction of the novelty based hypothesis, that hypothesis can't explain our model's first prediction. On the other hand, while the information gap and learning progress hypothesis were in line with our model's first prediction, both of these theories fail to explain our model's second prediction.   
 
\section{Testing the model predictions}

This section details the behavioral experiment that was conducted in order to test our model predictions. The experiment used two different scenarios -- confidence sampling and uniform sampling -- to assess whether people's curiosity is affected by changes in the relationship between need probability and confidence. In the confidence sampling condition, we created an environment such that need probability was related to confidence (Graph 1, Figure 1) and in the uniform sampling condition they were independent of each other (Graph 2, Figure 1). Based on our model predictions, we hypothesize that an inverted U-shape relation will exist between confidence and curiosity in the confidence condition and a decreasing relation will exist in uniform sampling condition. 

\subsection{Participants}

We recruited 298 participants from Amazon Mechanical Turk. They earned \$1.50 for participation with the option of earning an additional bonus of \$0.80. Participants in the experiment were randomly assigned to one of two conditions: confidence sampling condition (163 participants) and uniform sampling condition (135 participants). Informed consent was obtained using a consent form approved by the institutional review board at Berkeley.

\subsection{Stimuli}

The stimuli used in the experiment were 40 trivia questions on various topics that were taken directly from Experiment 1 in Kang et al. (2009). According to the authors, these questions were designed to measure curiosity about semantic knowledge and evoke a range of curiosity levels. 

\subsection{Procedure}

The experiment was divided into two phases -- the main round and bonus round. The main round was used to elicit and measure curiosity in participants. Participants were shown 40 trivia questions one after another and were asked to rate their confidence (i.e., probability that they know the correct answer) and curiosity in knowing the correct answer. Curiosity ratings were on a scale from 1 to 7 and the confidence scale ranged from 0 to 100\%. Following Kang et al.'s methodology, the raw curiosity ratings were individually normalized and confidence was rescaled to range from 0 to 1. The order of trivia questions was randomized for each participant. Thus, the main round of the experiment followed the procedure of Kang et al.'s design closely. This part of the experiment took approximately 7-8 minutes to complete. 

After the main round, the bonus round began which consisted of two parts. In the first part, all 40 questions from the main round were shown one after another and participants could choose to reveal the answer to those questions. However, each time they chose to reveal an answer, they had to wait an extra 10 seconds for the next question to appear. Findings from Experiment 3 of Kang et al. (2009) showed that participants were more likely to spend time, to wait longer, for the answers that they were more curious about. Thus, requiring participants to spend time to obtain information served as a proxy to measure their curiosity.

In the second part, participants attempted to answer 10 questions that were sampled from the main round (\$0.08 bonus for each correct answer). To discourse participants from using Google or other search engines, they were only given 2 minutes in total to answer the questions.

At the beginning of the experiment, participants were randomly assigned to two conditions -- the \textit{confidence} and the \textit{uniform} condition. Both the conditions had the same main round as described above but used different sampling methods for the bonus round. In the confidence condition, the sampling in bonus round was done based on the confidence ratings provided by the participants i.e. the questions for which participant's confidence rating was higher were more likely to appear in the second part of the bonus round. In the uniform condition, this sampling procedure was completely random i.e. each question was equally likely to appear in second part of the bonus round. Critically, participants were informed about the sampling procedure for their respective condition before the beginning of the bonus round. In a sense, the confidence condition creates a situation in which confidence is related to need probability (Graph 1 in Figure 1) and the uniform condition breaks this relationship (Graph 2 in Figure 1). 

According to our model's predictions we should see an inverted U-shape between curiosity and confidence for both the conditions in the main round. However, the curiosity of participants (i.e. the answers they revealed) should be different for both conditions in the bonus round. For the confidence condition participants' probability of revealing an answer should be highest for questions which they were moderately confident about. On the other hand, in the uniform condition, participants should be most curious about questions for which they were least confident about. 

\section{Results}

\begin{figure}[t]
    \centering
    \includegraphics[scale = 0.34]{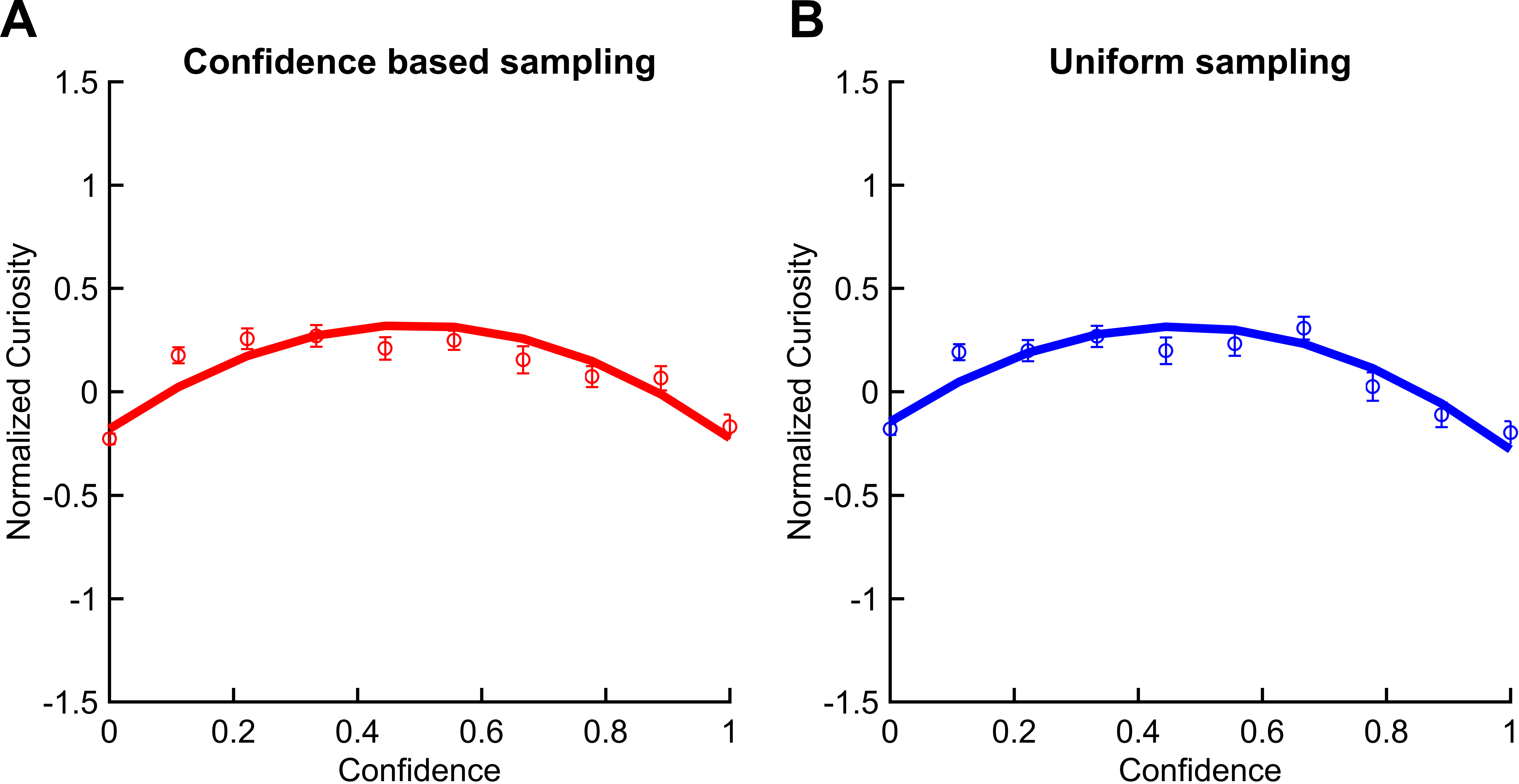}
    \caption{Relationship of curiosity and confidence in the main round for a) confidence condition and b) uniform condition. The markers indicate mean curiosity at each confidence level and the solid curve is the regression
    line. Curiosity is an inverted-U function of confidence for both conditions.}
\end{figure}

\begin{figure}[!b]
    \centering
    \includegraphics[scale = 0.135]{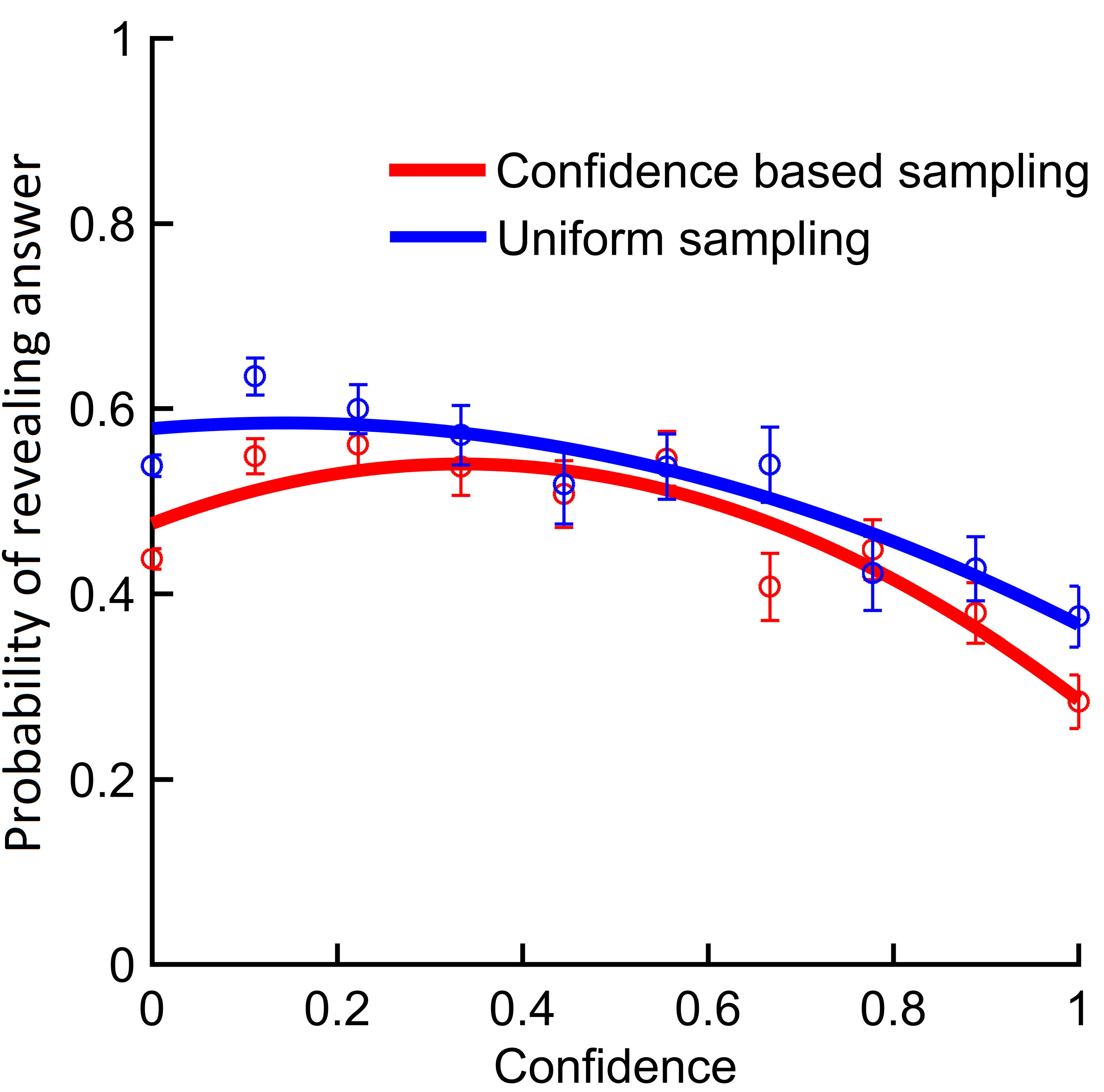}
    \caption{Probability of participants revealing an answer as a function of confidence in the bonus round. Consistent with our model's prediction, an inverted U-shape exists between curiosity and confidence in the confidence condition and a decreasing relationship exists in the uniform condition. }
\end{figure}

For all analyses that follow, we removed participants that revealed either too little ($<$3) or too many answers ($>$37) in the bonus round. 78 participants were excluded based on this criterion and our final data consisted of 220 participants (118 in the confidence condition and 102 in the uniform condition).

\textbf{Main Round. }Consistent with our prediction, an inverted U-shape exists between curiosity and confidence for both conditions (Figure 4). Following the method of Kang et al., we fitted curiosity with confidence and uncertainty i.e. confidence $\times$ (1-confidence) for both conditions. For the confidence condition, the model provided $r$ = 0.2 and a significant coefficient for uncertainty (estimate = 2.01, $p$ $<$ 0.001). For the uniform condition, the model provided similar results with $r$ = 0.2 and significant coefficient for uncertainty (estimate = 2.12, $p$ $<$ 0.001). Thus, for both conditions, the model provided a significant quadratic coefficient thereby demonstrating the prevalence of an inverted U-shape between curiosity and confidence for both conditions.

\textbf{Bonus Round.} We first computed the probability of participants revealing an answer conditioned on the confidence rating for both the conditions. As per our model's predictions, an inverted U-shape exists for the confidence condition and a decreasing relationship exists for the uniform condition (Figure 5). Similar to the previous analysis, we fitted confidence and uncertainty to both the conditions. For the confidence condition, the model provided $r$ = 0.9 and a significant coefficient for both confidence and uncertainty (estimates = -0.15 and 0.53 respectively with $p$ $<$ 0.05 for both) thereby showing a U-shape relationship. For the uniform condition, the model provided $r$ = 0.91 but the coefficient for uncertainty was not significant ($p$ = 0.09). On the other hand, the coefficient for confidence was significant (estimate = -0.23, $p$ $<$ 0.001), implying a decreasing relationship of curiosity with confidence for the uniform condition. 

\section{Discussion}

Curiosity is one of the hallmarks of human intelligence and is crucial to scientific discovery and invention. Models of curiosity have previously explained human curiosity by linking it to various psychological factors such as novelty, information-gap, and learning progress. We have shown that these different models are all special cases of curiosity | depending on the environment, curiosity can be driven by any of these factors.  Along with providing a way to unify previous distinct mechanistic accounts of curiosity, our rational model explains human curiosity in various settings.

Our results suggest that human curiosity is not only sensitive to the properties of the stimuli but it is also affected by the nature of the environment. If people are in an environment where need probability influences exposure, then their curiosity is highest for stimuli for which they are moderately confident about. On the other hand, if need probability and exposure are independent of each other then curiosity is highest for novel stimuli, i.e. stimuli for which people have little confidence. This can have important implications in the context of education where researchers are concerned with ways to pique curiosity in students. If we want to make people curious about tasks or activities for which they have little confidence in, perhaps subtle changes in the structure of the environment might be a step towards achieving that. We intend to explore such possibilities in future work, building upon the foundation established in this paper and working towards a better understanding of how to make people more curious especially in pedagogical settings.  

\nocite{anderson1990adaptive}
\nocite{baranes2015eye}
\nocite{berlyne1950novelty}
\nocite{berlyne1960conflict}
\nocite{duzel2010novelty}
\nocite{gobet2016bounded}
\nocite{gottlieb2013information}
\nocite{gruber2014states}
\nocite{kang2009wick}
\nocite{kidd2015psychology}
\nocite{law2016curiosity}
\nocite{loewenstein1994psychology}
\nocite{marr1982vision}
\nocite{oudeyer2006discovering}
\nocite{oudeyer2007intrinsic}
\nocite{ranganath2003neural}
\nocite{schmidhuber1991curious}
\nocite{schmidhuber2010formal}
\nocite{shannon}
\nocite{simon2001seek}
\nocite{smock1962children}
\nocite{Walin2016CuriosityAI}
\bibliographystyle{apacite}

\setlength{\bibleftmargin}{.125in}
\setlength{\bibindent}{-\bibleftmargin}
\def\bibfont{\small}
\bibliography{CogSci_Template}

\end{document}